%% file: main.tex
\documentclass[runningheads]{llncs}

\usepackage{eccv}

\input{preamble}
\usepackage{eccvabbrv}

\usepackage{graphicx}
\usepackage{booktabs}

\usepackage[accsupp]{axessibility}  

\usepackage[breaklinks,colorlinks,citecolor=eccvblue]{hyperref}

\usepackage{orcidlink}

\begin{document}

\title{SAGS: Structure-Aware 3D Gaussian Splatting} 


\author{Evangelos Ververas\inst{1,2}$^*$ \and
Rolandos Alexandros Potamias\inst{1,2}$^*$ \and
Jifei Song\inst{2} \and \\
Jiankang Deng\inst{1,2} \and
Stefanos Zafeiriou\inst{1} 
}

\authorrunning{E.~Ververas et al.}

\institute{Imperial College London \email{\{e.ververas16, r.potamias, j.deng16, s.zafeiriou\}@imperial.ac.uk}\\
\and
Huawei Noah's Ark Lab \email{\{jifeisong\}@huawei.com} \\
$^*$ Equal contribution}

\maketitle

\begin{abstract}
Following the advent of NeRFs, 3D Gaussian Splatting (3D-GS) has paved the way to real-time neural rendering overcoming the computational burden of volumetric methods. Following the pioneering work of 3D-GS, several methods have attempted to achieve compressible and high-fidelity performance alternatives. However, by employing a geometry-agnostic optimization scheme, these methods neglect the inherent 3D structure of the scene, thereby restricting the expressivity and the quality of the representation, resulting in various floating points and artifacts. In this work, we propose a structure-aware Gaussian Splatting method (SAGS) that implicitly encodes the geometry of the scene, which reflects to state-of-the-art rendering performance and reduced storage requirements on benchmark novel-view synthesis datasets. SAGS is founded on a local-global graph representation that facilitates the learning of complex scenes and enforces meaningful point displacements that preserve the scene's geometry. 
Additionally, we introduce a lightweight version of SAGS, using a simple yet effective mid-point interpolation scheme, which showcases a compact representation of the scene with up to 24$\times$ size reduction without the reliance on any compression strategies. Extensive experiments across multiple benchmark datasets demonstrate the superiority of SAGS compared to state-of-the-art 3D-GS methods under both rendering quality and model size. Besides, we demonstrate that our structure-aware method can effectively mitigate floating artifacts and irregular distortions of previous methods while obtaining precise depth maps.
Project page \url{https://eververas.github.io/SAGS/}.
\end{abstract}

\section{Introduction}
\label{sec:intro}
Novel View Synthesis (NVS) is a long-studied problem that aims to generate images of a scene from a specific point of view, using only a sparse set of images from different viewpoints with known camera parameters. Due to its diverse applications spanning from Virtual Reality (VR) \cite{deng2022fov} to content creation \cite{tang2023dreamgaussian,chen2023text}, novel view synthesis has garnered significant attention. With the advent of Neural Radiance Field (NeRF) \cite{mildenhall2021nerf}, an enormous amount of methods have been proposed to utilize volumetric rendering and learn implicit fields of the scene, achieving remarkable rendering results. However, albeit achieving highly detailed results, volumetric rendering methods fail to produce real-time renderings which hinders their real-world applications. 

\begin{figure}[t]
    \centering
    \includegraphics[width=\linewidth]{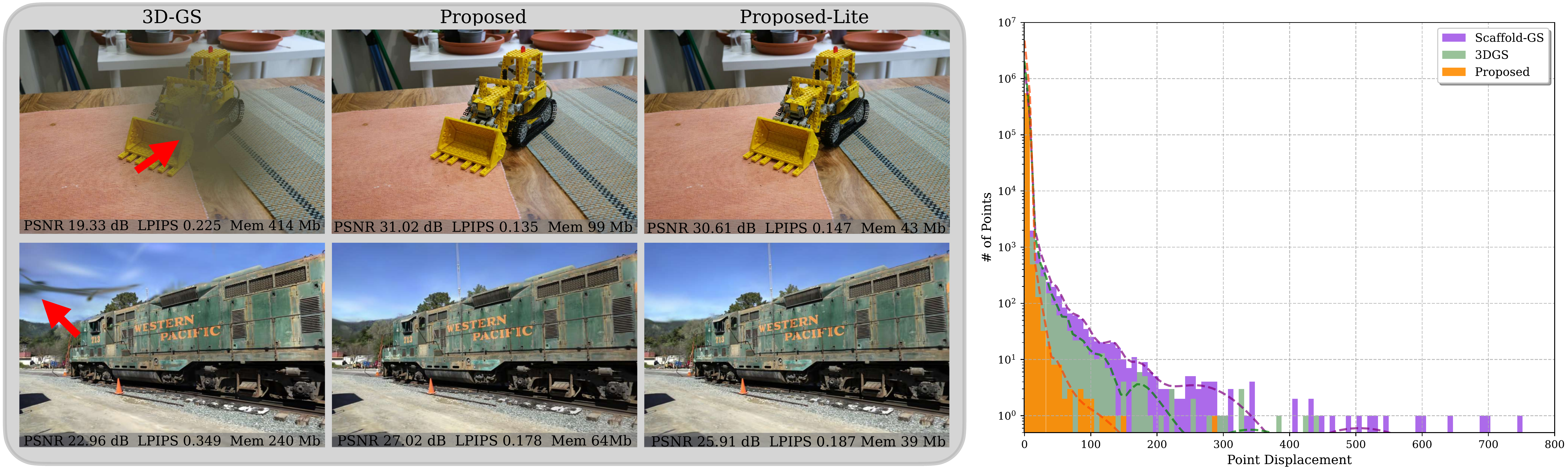}
    \caption{\textbf{Structure-Aware GS (SAGS)} leverages the intrinsic structure of the scene and enforces point interaction using graph neural networks outperforming the structure agnostic optimization scheme of 3D-GS \cite{kerbl3Dgaussians}. The 3D-GS method optimizes each Gaussian independently which results in 3D floaters and large point displacements from their original position (left). This can be also validated in the histogram of displacements (right) between the initial and the final position (mean) of a 3D Gaussian. Optimization-based methods neglect the scene structure and displace points far from their initial position to minimize rendering loss, in contrast to SAGS that predicts displacements that preserve the initial structure. The 3D-GS figures are taken directly from the original 3D-GS website.}
    \label{fig:displacements}
\end{figure}

Recently, Kerbl \etal \cite{kerbl3Dgaussians} introduced 3D Gaussian Splatting (3D-GS) to tackle this limitation using a set of differentiable 3D Gaussians that can achieve state-of-the-art rendering quality and real-time speed on a single GPU, outperforming previous NeRF-based methods \cite{mildenhall2021nerf,Barron2021MipNeRFAM,barron2022mipnerf360,chen2022mobilenerf}. In 3D-GS, the scene is parametrised using a set of 3D Gaussians with learnable shape and appearance attributes, optimized using differentiable rendering. To initialize the 3D Gaussians, Kerbl \etal \cite{kerbl3Dgaussians} relied on the point clouds derived from COLMAP \cite{sfm}, neglecting any additional scene structure and geometry during optimization.  

Undoubtedly, one of the primary drawbacks of the 3D-GS method is the excessive number of points needed to produce high-quality scene renderings. Following the success of 3D-GS, numerous methods have been proposed to reduce the storage requirements using compression and quantization schemes \cite{lee2023compact,niedermayr2023compressed,fan2023lightgaussian} while retaining the rendering performance. However, similar to 3D-GS, each Gaussian is optimized independently to fit the ground truth views, without the use of any additional structural inductive biases to guide the optimization. 
As can be seen in \cref{fig:displacements} (right), such a naive setting results in points being displaced far away from their initialization, thus neglecting their initial point structure and introducing floating points and artifacts \cite{xiong2023sparsegs} (highlighted in \cref{fig:displacements} (left) with red arrows). Apart from a significant degradation in the rendering quality, neglecting the scene’s geometry directly influences the scene's properties, including depth, which thereby limits its VR/AR applications. 

In this study, we propose a structure-aware Gaussian splatting method that aims to implicitly encode the scene's geometry and learn inductive biases that lead to point displacements that maintain the topology of the scene. Intuitively, points within the same local region often share common attributes and features, such as normals and color, that are neglected by current 3D-GS methods. Inspired by the success of Point Cloud analysis \cite{qi2017pointnet++}, we found our method on a graph constructed from the input scene and learn to model 3D Gaussians, as displacements from their original positions. The constructed graph serves as an inductive bias that aims to encode and preserve the scene structure while learning robust Gaussian attributes. Unlike the 3D-GS, the proposed method leverages the inter- and intra-connectivity between 3D point positions and learns to predict the Gaussian attributes using graph neural networks. Using both local and global structural information the network can not only reduce artifact floaters, but it can also increase the expressivity of the network leading to accurate scene representations compared to structure agnostic 3D-GS methods.

Under a series of experiments on different datasets and scenes, we evaluate the proposed method in terms of rendering quality, structure preservation, storage, and rendering performance, demonstrating the importance of structure in 3D Gaussian splatting. The proposed method can outperform the rendering quality of 3D-GS \cite{kerbl3Dgaussians} while 
reducing the memory requirements without sacrificing rendering speed. To sum up, our contributions can be summarized as follows: 
\begin{itemize}
    \item We introduce the first structure-aware 3D Gaussian Splatting method that leverages both local and global structure of the scene. 
    \item We bridge the two worlds between point cloud analysis and 3D Gaussian splatting, leveraging graph neural networks on the 3D point space. 
    \item We showcase that our method can produce state-of-the-art rendering quality, while reducing the memory requirements by up to 11.7$\times$ and 24$\times$ with our full and lightweight models, respectively.
\end{itemize}

\section{Related Work}
\label{sec:rel}
\noindent\textbf{Traditional scene reconstruction.}
Traditional 3D scene reconstruction techniques \cite{schonberger2016structure} utilize the structure-from-motion (SfM) pipeline \cite{sfm} for sparse point cloud estimation and camera calibration, and further apply multi-view stereo (MVS) \cite{goesele2007multi} to obtain mesh-based 3D scene reconstructions. Specifically, the traditional pipeline starts with a feature extraction step that obtains robust feature descriptors like \cite{lowe2004distinctive} and Superpoint \cite{detone2018superpoint}, followed by a feature matching module, \eg, SuperGlue \cite{sarlin2020superglue}, that matches the 2D image descriptors. After that, pose estimation and bundle adjustment steps are conducted to obtain all the reconstructed parameters, according to Incremental SfM \cite{cui2017batched}, Global SfM \cite{zhuang2018baseline}, or Hybrid-SfM \cite{cui2017hsfm}. Finally, MVS methods \cite{kaya2022uncertainty} are employed to reconstruct depth and normals of the target 3D object and subsequently fuse them to produce the final reconstruction. 

\smallskip
\noindent\textbf{NeRF based scene reconstruction.}
Neural radiance fields (NeRF) \cite{mildenhall2021nerf} introduced an implicit neural representation of 3D scenes, that revolutionized neural rendering based novel-view synthesis achieving remarkable photo-realistic renders. 
Several methods have extended NeRF model by including a set of appearance embeddings  \cite{martinbrualla2020nerfw} and improved training strategies \cite{tancik2022block,xiangli2022bungeenerf} to tackle complex and large-scale scenes. MipNeRF360 \cite{barron2022mipnerf360} achieved state-of-the-art rendering quality by addressing the aliasing artifacts observed in previous methods, suffering however from exceptionally slow inference times.
To improve the training and rendering efficiency of NeRFs, numerous methods have utilized grid-based structures to store compact feature representations. Interestingly, Plenoxels \cite{yu2022plenoxels} optimized a sparse voxel grid and achieved high-quality rendering performance without resorting to MLPs. Muller \etal \cite{muller2022instant} highlighted the importance of positional encodings for high-fidelity neural rendering and introduced a set of hash-grid encodings that significantly improved the expressivity of the model. 
Despite improving the efficiency compared to the global MLP representations, grid-based methods still struggle to achieve real-time rendering performance.

\smallskip
\noindent\textbf{3D-GS based scene reconstruction.}
Recently, 3D Gaussian splatting \cite{kerbl3Dgaussians} has been proposed to construct anisotropic 3D Gaussians as primitives, enabling high-quality and real-time free-view rendering. 
Similar to NeRFs, 3D-GS attempts to overfit the scene by optimizing the Gaussian properties. However, this usually results in an enormous amount of redundant Gaussians that hinder the rendering efficiency and significantly increase the memory requirements of the model. 
Several methods \cite{lee2023compact,niedermayr2023compressed} have attempted to reduce memory consumption by compressing the 3D Gaussians with codebooks, yet with the structure neglected, which could be vital in both synthesizing high-quality rendering and reducing the model size. Most relevant to our work, Scaffold-GS \cite{lu2023scaffold} introduced a structured dual-layered hierarchical scene representation to constrain the distribution of the 3D Gaussian primitives. However, Scaffold-GS still relies on structure-agnostic optimization which neglects the scene's global and local geometry, resulting in locally incoherent 3D Gaussians. This not only degrades the rendering quality but also significantly impacts the structural properties of the scene, including its depth. To tackle this, we devise a structure-aware Gaussian splatting pipeline that implicitly encodes the scene structure leveraging the inter-connectivity of the Gaussians. The proposed method introduces an inductive bias that not only facilitates high-fidelity renderings using more compact scene representations but also preserves the scene topology. 

\section{Method}
\subsection{Preliminaries: 3D Gaussian Splatting}
3D Gaussian Splatting \cite{kerbl3Dgaussians} is a state-of-the-art novel-view synthesis method that relies on explicit point-based representation. In contrast to the implicit representation of NeRFs that require computationally intensive volumetric rendering, 3D-GS renders images using the efficient point-based splatting rendering \cite{yifan2019differentiable,Wiles2019SynSinEV}. Each of the 3D Gaussians is parametrized by its position $\boldsymbol{\mu}$, covariance matrix $\boldsymbol\Sigma$, opacity $\boldsymbol\alpha$ and color $\boldsymbol c$ and has a density function defined as: 
\begin{equation}
    f(\boldsymbol{x}|\boldsymbol{\mu}, \boldsymbol{\Sigma}) = e ^ {- \frac{1}{2}(\boldsymbol{x}-\boldsymbol{\mu})^T \boldsymbol{\Sigma^{-1}} (\boldsymbol{x}-\boldsymbol{\mu})}
\end{equation}
where $\boldsymbol{x}$ is an arbitrary 3D point. 
To prevent the occurrence of non-positive semi-definite covariance matrices the authors proposed to decompose covariance matrix using rotation and scaling $\boldsymbol{\Sigma} = \mathbf{RSS^TR^T}$ where the rotation is represented using quaternions. Finally, the 3D Gaussians are splatted on the 2D image, and their corresponding 2D pixel color is calculated by blending the $N$ ordered Gaussians at the queried pixel as: 
\begin{equation}
    \mathbf{C} = \sum^{N}_{i=1} \alpha_i \prod_{j=1}^{i-1} (1 - \alpha_j) \mathbf{c}_i
\end{equation}
where $\mathbf{c}_i$ and $\alpha_i$ are the color and the opacity of Gaussian $i$. 

\subsection{Structure-Aware 3D Gaussian Splatting}
\begin{figure}[!t]
    \centering
    \includegraphics[width=\linewidth]{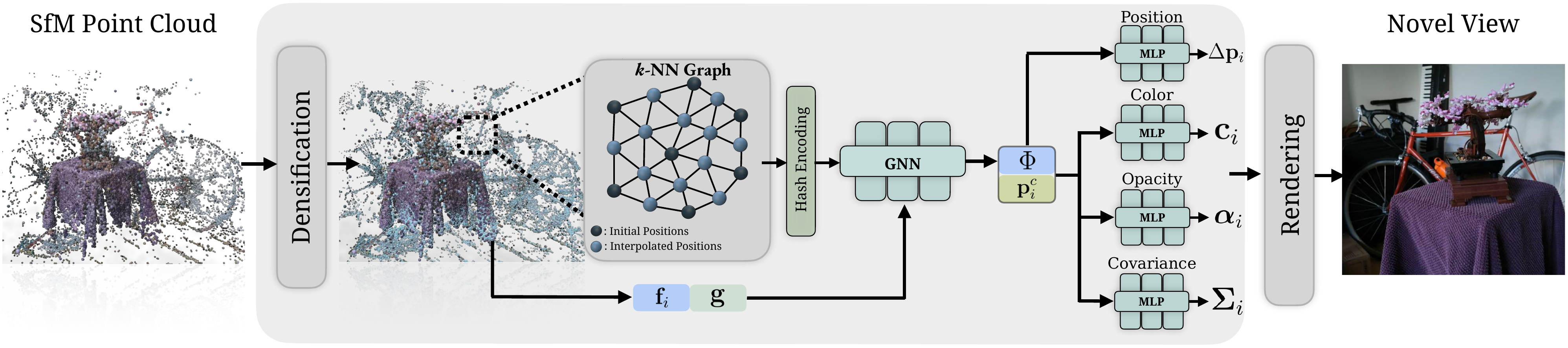}
    \captionof{figure}{\small{\textbf{Overview of the proposed method.} Given a point cloud obtained from COLMAP \cite{sfm}, we initially apply a curvature-based densification step to populate under-represented areas. We then apply $k$-NN search to link points $\mathbf{p}$ within local regions and create a point set graph. Leveraging the inductive biases of graph neural networks, we learn a local-global structural feature for each point $\Phi(\mathbf{p}_i, \mathbf{f}_i)$. Using a set of small MLPs we decode the structural features to 3D Gaussian attributes, i.e., color $\boldsymbol{c}$, opacity $\boldsymbol\alpha$, covariance $\boldsymbol{\Sigma}$ and point displacements $\Delta\mathbf{p}$ for the initial point position. Finally, we render the 3D Gaussians following the 3D-GS Gaussian rasterizer \cite{kerbl3Dgaussians}.} 
    \label{fig:Method}}
\end{figure}

In this work, we propose a structure-aware 3D Gaussian Splatting method, that takes as input a sparse point cloud $\mathbf{P} \in \mathbb{R} ^{M\times3}$ from COLMAP \cite{sfm} along with a set of sparse views with known camera parameters and learns a set of 3D Gaussians that fit the input views while maintaining the initial structure. The proposed method can be divided into three main components: a) the curvature-aware densification, b) the structure-aware encoder, and c) the refinement layer. Fig.~\ref{fig:Method} illustrates the pipeline of the proposed method. 

\subsubsection{Curvature-Aware Densification.} Undoubtedly, the performance of 3D-GS methods is significantly impacted by the sparse initialization of the Gaussians, which relies on COLMAP. In scenarios with challenging environments featuring texture-less surfaces, conventional Structure-from-Motion (SfM) techniques frequently fall short of accurately capturing the scene details, and thus are unable to establish a solid 3D-GS initialization. To tackle such cases, we introduce a densification step that aims to populate areas with zero or few points. 

\begin{figure}[!ht]
    \centering
    \includegraphics[width=0.8\linewidth]{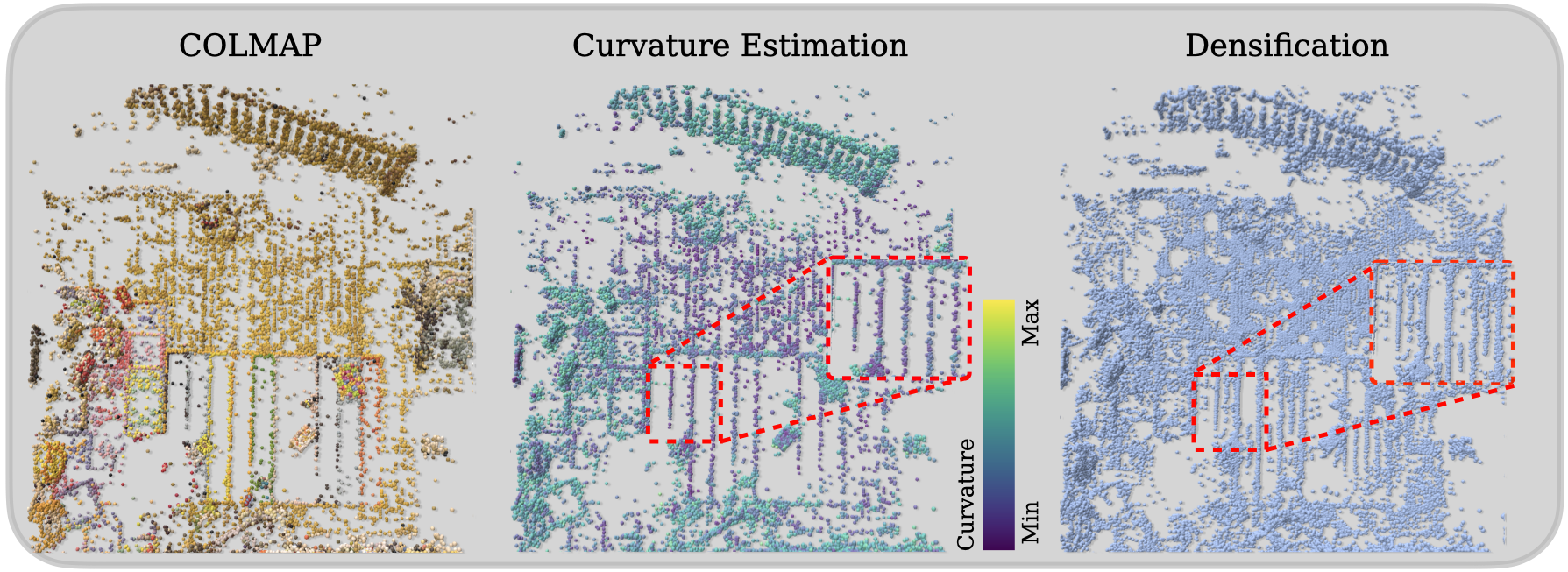}
    \captionof{figure}{\small{\textbf{Overview of the densification.} Given an initial SfM \cite{sfm} point cloud (left) we estimate the curvature following \cite{pauly2002efficient}. Curvature values are presented color-coded on the input COLMAP point cloud (middle) where colors with minimum curvature are closer to the purple color. The curvature-aware densification results in more points populating the low-curvature areas (right). } 
    \label{fig:densification}}
\end{figure}

In essence, 3D-GS methods attempt to reconstruct a scene from a sparse point cloud by employing a progressive growing scheme. This approach closely aligns with the well-explored field of point cloud upsampling. Drawing inspiration from the recent Grad-PU method \cite{He_2023_CVPR}, we incorporate an upsampling step to augment the point cloud's density, particularly in regions characterized by low curvature, which are typically under-represented in the initial COLMAP point cloud. We estimate the Gaussian mean curvature of the point cloud $\mathbf{P}$ following the local-PCA approach \cite{pauly2002efficient}. To generate a set of midpoints $\mathbf{p}_m$, we define a k-nearest neighbour graph for each of the low curvature points $\mathbf{p} \in \mathbf{P}_L \subset \mathbf{P}$ and calculate its midpoints as: 
\begin{equation}
    \mathbf{p}_m = \frac{1}{2} (\mathbf{p} + \mathbf{p}_j), \quad j \in k\text{-NN}(\mathbf{p})
\end{equation}
where $\mathbf{P}_L$ is the set of points $\mathbf{p}$ with curvature lower than a threshold and $\mathbf{p}_j$ is a neighbour of $\mathbf{p}$. An illustration of our densification approach is shown in \cref{fig:densification}.

Leveraging the mid-point densification step, we can train a lightweight model solely on the initial point set $\mathbf{P}$, while the remaining points, along with their attributes, can be defined on-the-fly.
This approach allows us to achieve both good performance and an extremely compact size, without the need for any compression scheme. We will refer to this model as \textbf{SAGS-Lite}. 

\subsubsection{Structure-Aware Encoder.}
Intuitively, points that belong to adjacent regions will share meaningful structural features that could improve scene understanding and reduce the floating artifacts. To enable point interactions within local regions and learn structural-aware features, we founded our method on a graph neural network encoder that aggregates local and global information within the scene. 

In particular, the first step of the proposed structure-aware network involves creating a $k$-Nearest Neighbour (NN) graph that links points within a local region. Using such $k$-NN graph we can enable point interaction and aggregate local features using graph neural networks. To further enhance our encoder with global structural information of the scene, we included a global feature that is shared across the points. In particular, the structure-aware GNN learns a feature encoding $\mathbf{f}_i$ for each point $\mathbf{p}_i \in \mathbf{P}$, using the following aggregation function: 
\begin{equation}
    \Phi(\mathbf{p}_i, \mathbf{f}_i) = \phi \left( \sum_{j \in \mathcal{N}(i)} w_{ij} h_{\mathbf{\Theta}} ( \gamma(\mathbf{p}_j), \mathbf{f}_j - \mathbf{f}_i, \mathbf{g}) \right)
\label{eq:gnn}
\end{equation}
where $\gamma(\cdot)$ denotes a positional encoding function that maps point $\mathbf{p}$ to a higher dimensional space $D$, $\mathbf{g}$ represents a global feature of the scene calculated as the maximum of the feature encoding $\text{max}(\mathbf{f})$, $h_{\mathbf{\Theta}}$ is a learnable MLP, $\phi$ represents a non-linear activation function and $w_{ij}$ is an inverse-distance weight defined from the softmax of the inverse distances between the point $\mathbf{p}_i$ and its neighbors $\mathcal{N}(i)$:
\begin{equation}
    w_{ij} = \frac{\text{dist}^{-1}(\mathbf{p}_i, \mathbf{p}_j)}{\sum_{j \in \mathcal{N}(i)}\text{dist}^{-1}(\mathbf{p}_i, \mathbf{p}_j)}
\end{equation}
Following \cite{wang2019dynamic,li2019deepgcns}, we opted to utilize relative features $\mathbf{f}_j - \mathbf{f}_i$ since it is more efficient than aggregating raw neighborhood
features and it enriches the expressivity of the network. To encode Gaussian positions we selected the high-performing multi-resolution hash encoding \cite{muller2022instant} given its lightweight nature and its ability to expressively encode complex scenes.

\subsubsection{Refinement Network.}
In the final state of the proposed model, the structure-aware point encodings are decoded to the 3D Gaussian attributes using four distinct networks, one for each of the attributes; namely position $\boldsymbol{\mu} \in \mathbb{R}^3$, color $\mathbf{c} \in \mathbb{R}^3$, opacity $\boldsymbol{\alpha} \in \mathbb{R}^1$ and covariance $\boldsymbol{\Sigma} \in \mathbb{R}^{3\times3}$. Aligned with 3DGS, we parametrize covariance matrix $\boldsymbol{\Sigma}$, with a scale vector $\mathbf{S} \in \mathbb{R}^3$ and a rotation matrix $\mathbf{R}\in \mathbb{R}^{3\times3}$ represented with quaternions. To enforce high rendering speed, we defined each decoder as a small MLP that takes as input the structure-aware encoding and the view-dependent point positions $\mathbf{p}_i$ and outputs the Gaussian attributes for each point. For example, the color attribute $\mathbf{c}$ can be defined as:
\begin{equation}
    \mathbf{c}_i = \text{MLP}_{\mathbf{c}}(\Phi(\mathbf{p}_i, \mathbf{f}_i), \mathbf{p}^{\mathbf{c}}_i)
\end{equation}
where $\text{MLP}_{\mathbf{c}}(\cdot)$ represents the color attribute MLP layer and $\mathbf{p}^{\mathbf{c}}_i$ are the view-depended point positions, normalized with the camera coordinates $\mathbf{x}_c$ as: 
\begin{equation}
   \mathbf{p}^{\mathbf{c}}_i  = \frac{\mathbf{p}_i - \mathbf{x}_c}{||\mathbf{p}_i - \mathbf{x}_c||_2}
\end{equation}
Similarly, we predict opacity $\boldsymbol{\alpha}$, scale $\boldsymbol{S}$ and rotation $\boldsymbol{R}$ attributes using view-depended point positions.
In contrast to the aforementioned view-dependent Gaussian attributes, we opted to learn the 3D scene, represented from the Gaussian mean positions $\boldsymbol{\mu}$, in a camera-agnostic manner. This way we can enforce the model to learn the underlying 3D geometry solely using the world-space point position and shift the bulk of the view-depended attributes to the rest of the MLPs.
Additionally, to enforce stable training, we model the 3D Gaussian positions $\boldsymbol{\mu}$ as displacement vectors from the initial COLMAP positions: 
\begin{equation}
    \boldsymbol{\mu}_i = \mathbf{p}_i + \Delta \mathbf{p}_i,  \quad \Delta \mathbf{p}_i = \text{MLP}(\Phi(\mathbf{p}_i, \mathbf{f}_i))
\end{equation}
where $\mathbf{p}_i$ denotes the initial point derived from structure-from-motion. 

\subsubsection{SAGS-Lite.} In addition to our best-performing model, we present a simple but effective strategy to reduce the storage requirements of the model while retaining the high quality of the renderings and its structural properties. Considering that the predominant storage burden in 3D Gaussian Splatting methods stems from the abundance of stored Gaussians, our objective was to devise a pipeline that would yield a much more compact set of Gaussians, without relying on vector quantization or compression techniques. In particular, leveraging the initial densification step, we can define the midpoints using the initial key points of the COLMAP and predict their Gaussian attributes using the interpolated key-point features.

Under this setting, the mid-points can be generated on the fly and their corresponding features will be interpolated from the structure-aware feature encodings $\mathbf{f}'$ as: 
\begin{equation}
    \mathbf{f}'_m = \frac{1}{2}(\mathbf{f}'_i + \mathbf{f}'_j), \quad (i,j) \in \mathbf{P}
\end{equation}
where $\mathbf{f}'_i, \mathbf{f}'_j$ are the feature encodings of two keypoints $i,j \in \mathbf{P}$ and $\mathbf{f}'_m$ defines the interpolated feature of their midpoint. Aligned with our full model, the interpolated features along with their corresponding view-depended interpolated positions are fed to the refinement networks to predict their Gaussian attributes. 

\noindent\textbf{Training.}
To train our model we utilized a $\mathcal{L}_1$ loss and a structural-similarity loss $\mathcal{L}_{SSIM}$ on the rendered images, following \cite{kerbl3Dgaussians}: 
\begin{equation}
    \mathcal{L} = (1-\lambda)\mathcal{L}_1 + \lambda \mathcal{L}_{SSIM}
\end{equation}
where $\lambda$ is set to 0.2. 

\noindent\textbf{Implementation.}
We build our model on top of the original 3D-GS~\cite{kerbl3Dgaussians} PyTorch implementation. Similar to 3D-GS, we train our method for 30,000 iterations across all scenes and apply a growing and pruning step until 15,000 iterations, every 100 iterations, starting from the 1500 iterations. Throughout our implementation, we utilize small MLPs with a hidden size of 32.

\section{Experiments}
\textbf{Datasets.} To evaluate the proposed method, on par with the 3D-GS \cite{kerbl3Dgaussians}, we utilized 13 scenes including nine scenes from Mip-NeRF360 \cite{barron2022mipnerf360}, two scenes from Tanks\&Temples \cite{Knapitsch2017} and two scenes from Deep Blending \cite{DeepBlending2018} datasets.

\noindent\textbf{Baselines.} We compared the proposed method with NeRF- and 3D-GS-based state-of-the-art works in novel-view synthesis, including the Mip-NeRF360 \cite{barron2022mipnerf360}, Plenoxels \cite{yu2022plenoxels}, iNGP \cite{muller2022instant}, 3D-GS \cite{kerbl3Dgaussians} along with the recent Scaffold-GS \cite{lu2023scaffold}. 

\noindent\textbf{Metrics.} We evaluate the proposed SAGS model in terms of rendering quality, structure preservation, and rendering performance. To measure the rendering quality, we utilized the commonly used PSNR, SSIM \cite{1284395}, and LPIPS \cite{Zhang_2018_CVPR} metrics. In addition, we report model storage requirements in megabytes (MB) and rendering speed in frames per second (FPS). 

\subsection{Novel-View Synthesis}
\begin{figure}[!ht]
    \centering
    \includegraphics[width=\linewidth]{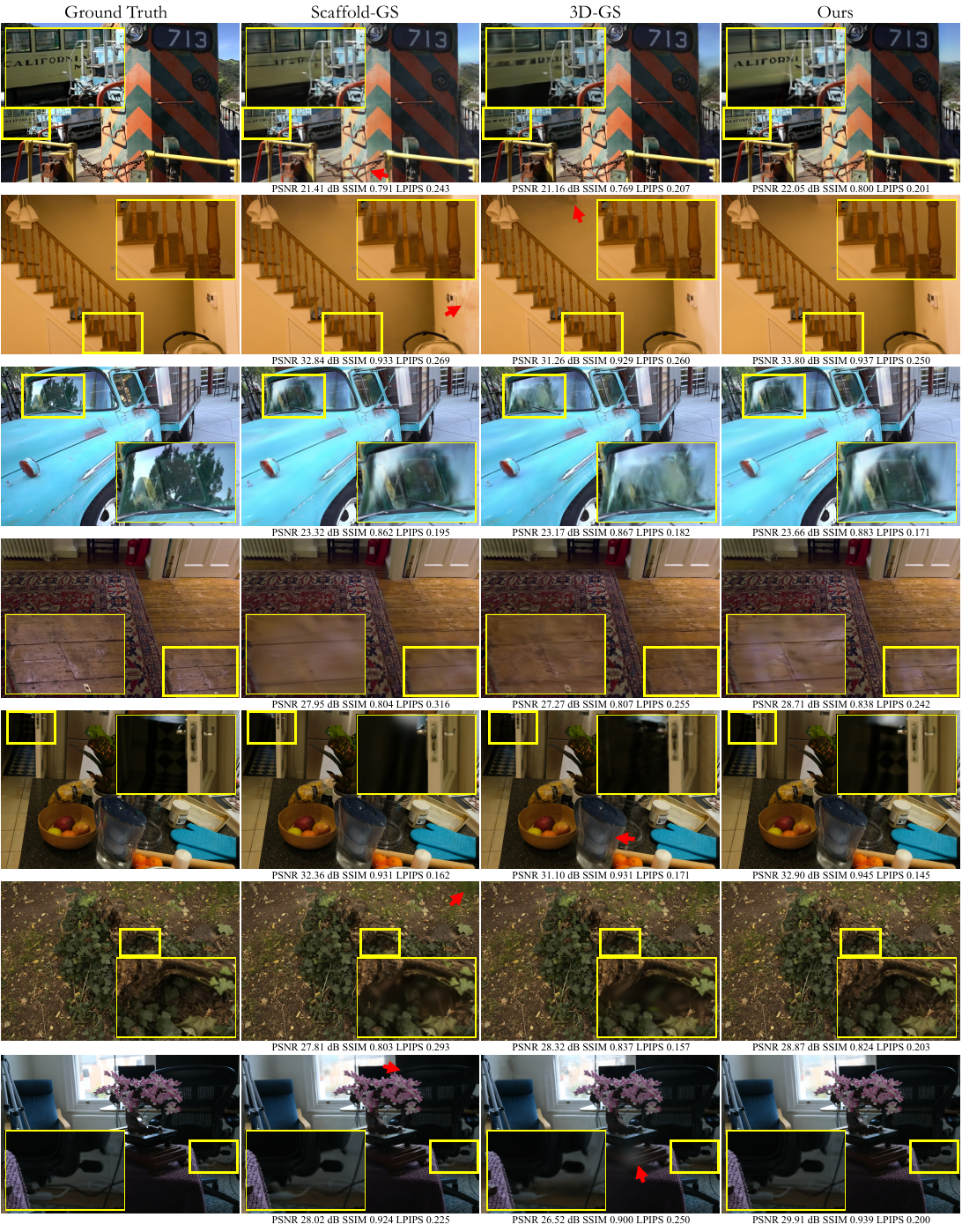}
    \captionof{figure}{\small{\textbf{Qualitative comparison.} We qualitatively evaluate the proposed and the baseline methods (3D-GS~\cite{kerbl3Dgaussians} and Scaffold-GS~\cite{lu2023scaffold}) across six scenes from different datasets. We highlight some detailed differences between the three methods using a \textcolor{yellow}{magnified crop} in yellow. We also emphasize additional visual artifacts using \textcolor{red}{red arrows}. The proposed method consistently captures more structural and high-frequency details while minimizing floaters and artifacts compared to the baseline methods.}
    \label{fig:sota}}
\end{figure}

\textbf{Rendering Quality.} In \cref{tab:performance}, we report the average evaluation performance of the proposed and the baseline methods over the three datasets. As can be easily seen, SAGS outperforms 3D-GS and the recently introduced Scaffold-GS method under all datasets and metrics. Leveraging the inter-connectivity between the 3D Gaussians, the SAGS model can facilitate high-quality reconstruction in challenging cases that the independent and unstructured optimization scheme of 3D-GS and Scaffold-GS methods struggle. As can be qualitatively validated in \cref{fig:sota}, the proposed SAGS model can better capture high-frequency details, such as the letters on the train wagon, the door handle, and the desk chair mechanism.

\input{Tables/tab_results}

\begin{figure}[!ht]
    \centering
    \includegraphics[width=0.8\linewidth]{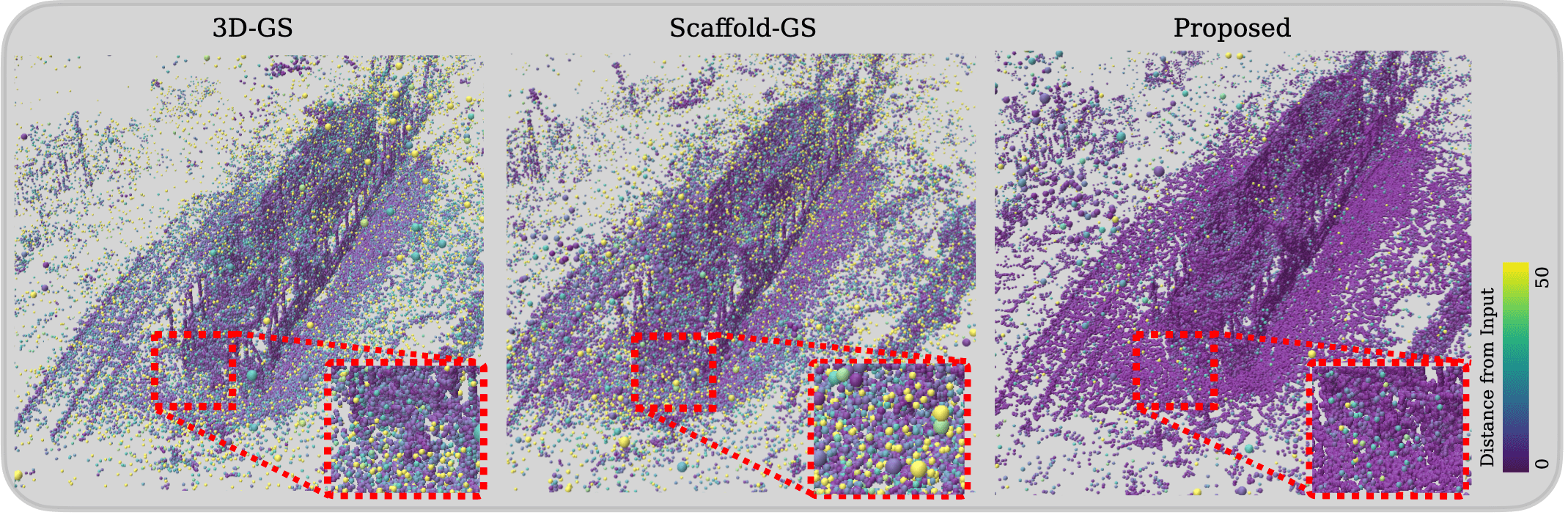}
    \captionof{figure}{\small{\textbf{Color Coded Gaussian Displacements.} We measured the Gaussians' displacements from their original positions, on the ``train'' scene from Tanks\&Temples \cite{Knapitsch2017} dataset, and encoded them in a colormap scale. Colors closer to purple color indicate small displacements. Both the 3D-GS and Scaffold-GS methodologies depend on a rudimentary point optimization approach, that neglects the local topology and fails to guide the Gaussians in a structured manner.}
    \label{fig:colorcoded}}
\end{figure}

\noindent \textbf{Structure Preservation.} Apart from the visual quality of the renderings, preserving the 3D geometry of the scene is extremely crucial for downstream VR/AR applications of NVS. Using the proposed structure-aware encoder, we manage to tackle the structure preservation limitations of previous 3D-GS methods and constrain the point displacements close to their initial positions. As pointed out with the red arrows in \cref{fig:sota}, this substantially diminishes floater artifacts, which were prevalent in the 3D-GS method. To quantitatively validate the structural preservation of our method, in \cref{fig:colorcoded}, we illustrate the displacements of points, in a color-coded format, on top of their original positions. In particular, we depict the color-coded displacements for the \emph{train} scene from the Tanks\&Temples dataset, where points with color closer to purple indicate small displacements and colors closer to yellow indicate large displacements. Aligned with the findings of \cref{fig:displacements} (right), SAGS better constrains the Gaussians to lie in the original geometry of the scene compared to 3D-GS and Scaffold-GS methods that rely on structure-agnostic optimization to fit the scene.    

\begin{figure}[!ht]
    \centering
    \includegraphics[width=\linewidth]{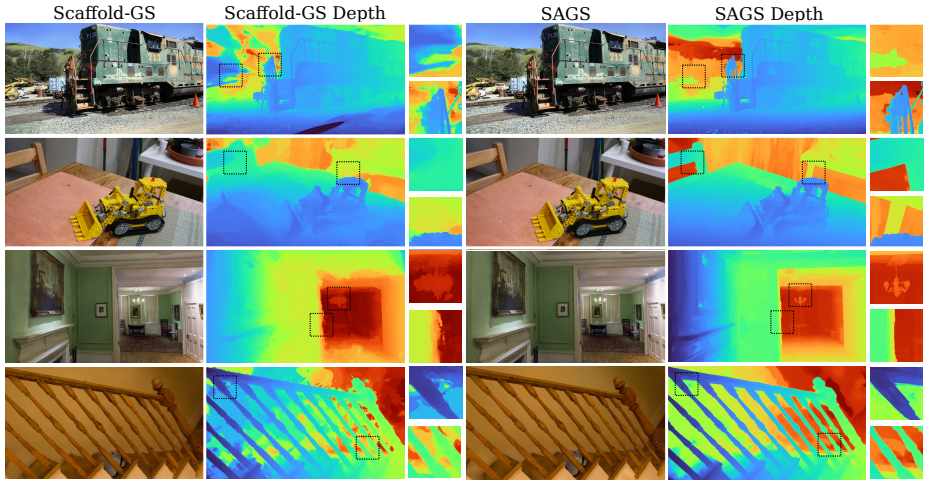}
    \captionof{figure}{\small{\textbf{Depth Structural Preservation.} Comparison between the proposed and the Scaffold-GS method on the scene's structure preservation. The proposed method can accurately capture sharp edges and suppress ``floater'' artifacts that are visible on the Scaffold-GS depth maps.}
    \label{fig:depth}}
\end{figure}

In addition, preserving the geometry of the scene ensures the preservation of spatial relationships and distances between objects, which is significantly important in 3D modeling. The depth information provides crucial cues that verify the spatial distances within a scene and can easily validate the suppression of floater artifacts that are not visible in the rendered image. Therefore, in \cref{fig:depth}, we qualitatively evaluate the depth maps that correspond to various scene renderings, generated by the SAGS and the Scaffold-GS methods. The proposed method can not only achieve sharp edges and corners (\eg, on the table and the door) but also accurate high-frequency details (\eg, the tire track of the bulldozer and the staircase banister). On the contrary, Scaffold-GS method provides noisy depth maps that fail to capture the scene's geometry, \eg, the chair back and the chandelier are modeled with an enormous set of points that do not follow any 3D representation. This is caused by the unstructured nature of the Gaussian optimization that attempts to minimize only the rendering constraints without any structural guidance. Furthermore, Scaffold-GS method falls short in accurately representing flat surfaces, as can be seen in the walls and the table, compared to SAGS which accurately captured flat surfaces. 

\noindent\textbf{Performance.} Apart from the rendering quality we evaluated the performance of the proposed and the baseline methods in terms of rendering speed (FPS) and storage size (MB) under all datasets. In \cref{tab:size_fps}, we report the FPS and memory sizes for each method averaged per dataset. Both SAGS and SAGS-Lite models achieve a real-time rendering speed, with over 100 FPS under all scenes, on par with the Scaffold-GS and 3D-GS methods. Importantly, SAGS reduces the storage requirements of 3D-GS by 5$\times$ on the challenging MipNeRF360 dataset \cite{barron2022mipnerf360} achieving state-of-the-art performance with an average model of 135MB. The storage requirements are reduced even more with our lightweight model that can achieve up to 24$\times$-storage reduction compared to 3D-GS without relying on any compression scheme. 
\input{Tables/tab_size_fps}

\begin{figure}[!ht]
    \centering
    \includegraphics[width=\linewidth]{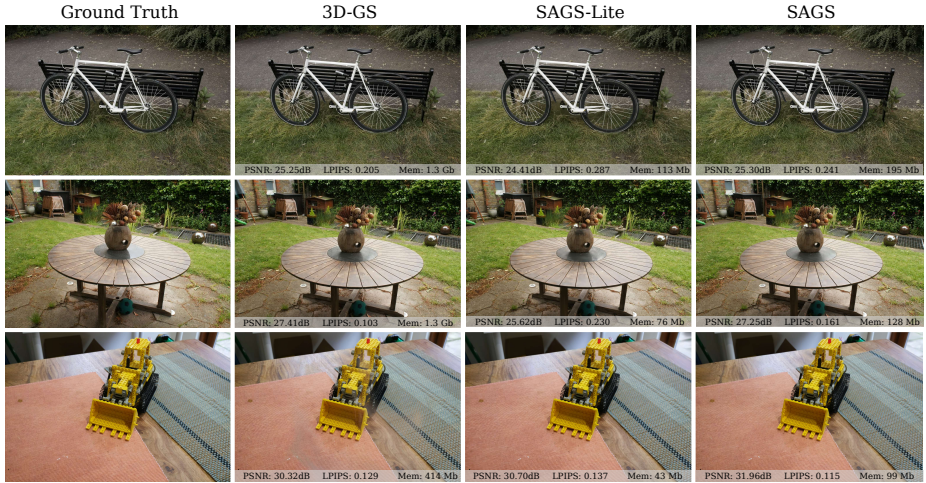}
    \captionof{figure}{\small{\textbf{Comparison with SAGS-Lite.} We qualitatively compared the proposed SAGS-Lite model against SAGS and 3DGS. SAGS-Lite can achieve to maintain high quality renderings with up to 24$\times$ storage reduction compared to 3DGS.  }
    \label{fig:sags_lite}}
\end{figure}

\noindent\textbf{Comparison with SAGS-Lite.} In \cref{fig:sags_lite}, we qualitatively evaluate the light-weight version of the proposed SAGS model. As shown in \cref{tab:size_fps}, the SAGS-Lite model can drastically reduce the storage requirements of 3D-GS by up to 28 times, while achieving similar rendering performance with 3D-GS \cite{kerbl3Dgaussians}. Despite lacking some sharp details when compared to our full SAGS model, SAGS-Lite can accurately represent the scene while at the same time avoid the floater artifacts caused by 3D-GS optimization.  



\subsection{Ablation Study}

\begin{figure}[!t]
    \centering
    \includegraphics[width=\linewidth]{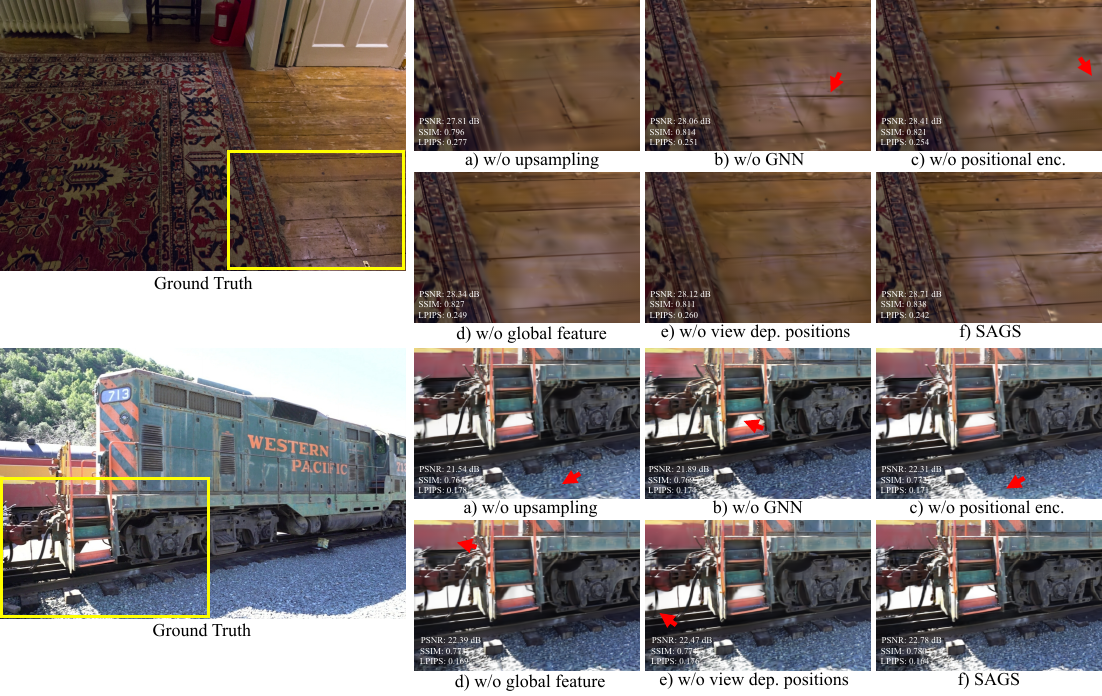}
    \captionof{figure}{\small{\textbf{Ablation study on the components of SAGS.} We perform a series of ablation experiments on the Deep Blending and the Tanks\&Temples datasets and demonstrate qualitative results from the \emph{drjohnson} and the \emph{train} scenes. We emphasise differences across model configurations over the same \textcolor{yellow}{crop} of the resulting images and highlight additional visual artifacts using \textcolor{red}{red} arrows.}}
    \label{fig:ablation}
\end{figure}

To evaluate the impact of individual components within the proposed method, we conducted a series of ablation studies using the Deep Blending and Tanks\&Temples datasets. 
The quantitative and qualitative results of these studies are presented in \cref{tab:ablation} and \cref{fig:ablation}, respectively. 

\noindent\textbf{Effect of the Densification.} First, removing the curvature-aware densification step leads to parts of the scene being under-represented in the resulting point cloud, as the gradient-based growing struggles to sufficiently fill them.
This is particularly evident in areas of COLMAP which are initially undersampled, such as the floor in the \emph{drjohnson} scene and the gravel road in the \emph{train} scene. 
In contrast, our curvature-aware densification adequately fills those areas supporting further growing during training.

\noindent\textbf{Effect of the Structure-Aware Encoder.} Replacing the aggregation layer in \cref{eq:gnn} with an MLP diminishes the inductive biases and the expressivity of the model to encode the scene's structure, leading to Gaussians with locally inconsistent attributes. As can be seen in \cref{fig:ablation}, the absence of structure results in renderings with more floaters and artifacts, which are evident in both the \emph{drjohnson} and \emph{train} scenes. 

\noindent\textbf{Feature Analysis.} Using the points' positions directly in \cref{eq:gnn} instead of their positional encodings $\gamma(\mathbf{p})$, results in lower resolution representations of the scene which implies less high frequency details in renderings.
A similar effect is caused by removing the global structure information offered by $\mathbf{g}$, which leads to less expressive feature encodings $\Phi(\mathbf{p}_i, \mathbf{f}_i)$ limiting the capacity of the refinement network and the quality of its predictions.
For both previous configurations, the floor in the \emph{drjohnson} scene and areas of the \emph{train} scene demonstrate parts with flat textures.
Last, by ablating the view dependent positions $\mathbf{p}_{i}^{c}$ from the appearance related attributes resulted in missing reflections, for example on the floor in \emph{drjohnson}, and floating points as on the black cable in the \emph{train} scene.

\input{Tables/tab_ablation}

\section{Conclusion}
In this paper, we present Structure-Aware Gaussian Splatting (SAGS), a novel Gaussian Splatting approach that leverages the intrinsic scene structure for high-fidelity neural rendering. Motivated by the shortcomings of current 3D Gaussian Splatting methods to naively optimize Gaussian attributes independently neglecting the underlying scene structure, we propose a graph neural network based approach that predicts Gaussian's attributes in a structured manner. Using the proposed graph representation, neighboring Gaussians can share and aggregate information facilitating scene rendering and the preservation of its geometry. We showcase that the proposed method can outperform current state-of-the-art methods in novel view synthesis while retaining the real-time rendering of 3D-GS. We further introduce a simple yet effective mid-point interpolation scheme that attains up to 24$\times$-storage reduction compared to 3D-GS method while retaining high-quality rendering, without the use of any compression and quantization algorithm. Overall, our findings demonstrate the benefits of introducing structure in 3D-GS. 
%
%
\bibliographystyle{splncs04}
\bibliography{main}

\end{document}

%% file: preamble.tex
\usepackage[dvipsnames]{xcolor}
\usepackage{colortbl}

\definecolor{placeholder}{rgb}{0.6,0.8,0.95}

\definecolor{amber}{rgb}{1.0, 0.75, 0.0}

\definecolor{visible-blue}{rgb}{0.286, 0.525, 0.910}

\definecolor{tabfirst}{rgb}{1, 0.7, 0.7} 
\definecolor{tabsecond}{rgb}{1, 0.85, 0.7} 
\definecolor{tabthird}{rgb}{1, 1, 0.7} 

%% file: Tables/tab_results.tex
\begin{table*}[]
\centering
\caption{\textbf{Quantitative comparison} between the proposed and the baseline methods on Mip-NeRF360 \cite{barron2022mipnerf360}, Tanks\&Temples \cite{Knapitsch2017} and Deep Blending \cite{DeepBlending2018} datasets.}
\label{tab:performance}
\resizebox{\linewidth}{!}{
\begin{tabular}{l|ccc|ccc|ccc}
\toprule
Dataset & \multicolumn{3}{c|}{Mip-NeRF360} & \multicolumn{3}{c|}{Tanks\&Temples} & \multicolumn{3}{c}{Deep Blending} \\
\begin{tabular}{c|c} Method & Metrics \end{tabular}  & PSNR \(\uparrow\) & SSIM \(\uparrow\) & LPIPS \(\downarrow\) & PSNR \(\uparrow\) & SSIM \(\uparrow\) & LPIPS \(\downarrow\) & PSNR \(\uparrow\) & SSIM \(\uparrow\) & LPIPS \(\downarrow\) \\
\midrule

3D-GS~\cite{kerbl3Dgaussians} & 
28.69 & \cellcolor{tabsecond}0.870 &\cellcolor{tabsecond} 0.182 & 
23.14 & 
0.841 & 0.183 & 
\cellcolor{tabthird}29.41 & \cellcolor{tabthird}0.903 & \cellcolor{tabsecond}0.243 \\

Mip-NeRF360 ~\cite{barron2022mipnerf360} & 
\cellcolor{tabsecond}29.23 & 0.844 & \cellcolor{tabthird}0.207 & 
22.22 & 0.759 & 0.257 & 
29.40 & 0.901 & \cellcolor{tabthird}0.245 \\

iNPG ~\cite{muller2022instant} & 26.43 & 0.725 & 0.339 & 21.72 & 0.723 & 0.330 & 23.62 & 0.797 & 0.423 \\

{Plenoxels} ~\cite{yu2022plenoxels} & 23.62 & 0.670 & 0.443 & 21.08 & 0.719 & 0.379 & 23.06 & 0.795 & 0.510 \\ 

Scaffold-GS ~\cite{lu2023scaffold} & 
\cellcolor{tabthird}28.84 & \cellcolor{tabthird}0.848 & 0.220 & 
\cellcolor{tabthird}23.96 & \cellcolor{tabsecond}0.853 & \cellcolor{tabsecond}0.177 & 
\cellcolor{tabsecond}30.21 & \cellcolor{tabsecond}0.906 & 0.254 \\
\hline
SAGS-Lite & 
    28.54 & 0.841 & 0.225 & 
    \cellcolor{tabsecond}24.16 & \cellcolor{tabthird}0.846 & \cellcolor{tabthird}0.181 & 
    29.07 & 0.889 & 0.292 \\
    
SAGS & 
    \cellcolor{tabfirst}29.65 & \cellcolor{tabfirst}0.874 & \cellcolor{tabfirst}0.179 & 
    \cellcolor{tabfirst}24.88 & \cellcolor{tabfirst}0.866 & \cellcolor{tabfirst}0.166 & 
    \cellcolor{tabfirst}30.47 & \cellcolor{tabfirst}0.913 & \cellcolor{tabfirst}0.241 \\
\bottomrule
\end{tabular}}
\end{table*}

%% file: Tables/tab_size_fps.tex
\begin{table}[t]
\centering
\caption{\textbf{Performance comparison} 
between the proposed and the 3D-GS based methods. We also report the storage reduction of each model compared to original 3D-GS method \cite{kerbl3Dgaussians}. 
}
\label{tab:size_fps}
\resizebox{0.7\linewidth}{!}{
\begin{tabular}{l|cc|cc|cc}
\toprule
Dataset & \multicolumn{2}{c|}{Mip-NeRF360} & \multicolumn{2}{c|}{Tanks\&Temples} & \multicolumn{2}{c}{Deep Blending}  \\
Methods & FPS & Mem (MB) & FPS & Mem (MB) & FPS & Mem (MB) \\ 
\midrule
3D-GS~\cite{kerbl3Dgaussians} & 97 & 693 & 123 & 411 & 109 & 676 \\
Scaffold-GS~\cite{lu2023scaffold}  & 102 & 252 (2.8$\times$ $\downarrow$) & 110 & 87 (4.7$\times$ $\downarrow$) & 139 & 66 (10.2$\times$ $\downarrow$) \\

Ours & \textbf{110} & 135 (5.1$\times$ $\downarrow$) & 108 & 75 (5.5$\times$ $\downarrow$) & 138 & 58 (11.7$\times$ $\downarrow$) \\
Ours-Lite & 101 & \textbf{76} (\textbf{9.1}$\times$ $\downarrow$) & \textbf{112} & \textbf{35} (\textbf{12}$\times$ $\downarrow$) & \textbf{142} & \textbf{28} (\textbf{24}$\times$ $\downarrow$) \\

\bottomrule

\end{tabular}

}
\end{table}

%% file: Tables/tab_ablation.tex
\begin{table*}[t]
\centering
\caption{\textbf{Ablation study on the components of SAGS.} The ablation was performed on the Deep Blending and the Tanks\&Temples datasets. 
}
\label{tab:ablation}
\resizebox{\linewidth}{!}{
\begin{tabular}{l|ccc|ccc}
\toprule

Scene & \multicolumn{3}{c|}{Deep Blending} & \multicolumn{3}{c}{Tanks\&Temples} \\

\begin{tabular}{c|c} Ablation & Metrics \end{tabular}  & PSNR \(\uparrow\) & SSIM \(\uparrow\) & LPIPS \(\downarrow\) & PSNR \(\uparrow\) & SSIM \(\uparrow\) & LPIPS \(\downarrow\) \\
\midrule

w/o Curvature-Aware Densification & 
29.87 & 0.901 & 0.259 & 
23.97 & 0.851 & 0.175 \\

w/o GNN & 
29.94 & 0.905 & 0.254 & 
24.19 & 0.844 & 0.181 \\

w/o Positional-Encoding $\gamma(\mathbf{p})$ & 
30.21 & 0.909 & 0.252 & 
24.31 & 0.852 & 0.169 \\

w/o Global Feature $\mathbf{g}$ & 
30.17 & 0.911 & 0.250 & 
24.42 & 0.861 & 0.174 \\

w/o View Dependent Positions $\mathbf{p}^{\mathbf{c}}_i$ & 
30.07 & 0.903 & 0.256 & 
24.37 & 0.849 & 0.173 \\

\hline
SAGS & 
    \textbf{30.47} & \textbf {0.913} & \textbf{0.241 }& 
   \textbf {24.88} & \textbf {0.866} & \textbf {0.166 }\\
\bottomrule
\end{tabular}}
\end{table*}